\RequirePackage{amsmath, accents}
\documentclass[runningheads]{llncs}
\usepackage[T1]{fontenc}
\usepackage{graphicx}
\usepackage{algorithmic}
\usepackage[linesnumbered,ruled]{algorithm2e}

\usepackage{xcolor}
\usepackage{bm}
\usepackage{scalerel}

\newcommand{\todo}[1]{{\color{red}{#1}}}
\usepackage{amsmath,amssymb,amsfonts, mathtools}
\DeclarePairedDelimiter\abs{\lvert}{\rvert}%
\DeclarePairedDelimiterX{\norm}[1]{\lVert}{\rVert}{#1}

\DeclareMathOperator*{\argmax}{argmax} 
\DeclareMathOperator*{\argmaxgreedy}{argmax\_greedy} 

\usepackage{accents}
\newcommand{\ubar}[1]{\underaccent{\bar}{#1}}

\let\oldnl\nl
\newcommand{\nonl}{\renewcommand{\nl}{\let\nl\oldnl}}%
\usepackage[caption=false]{subfig}

\usepackage[numbers,sort&compress]{natbib}  

\begin{document}
\title{Inverse Risk-sensitive Multi-Robot Task Allocation}
%
%
\author{Guangyao Shi \and
Gaurav S. Sukhatme }
\authorrunning{G. Shi et al.}
%
\institute{ University of Southern California, Los  Angeles CA  90007, USA \\
\email{shig, gaurav@usc.edu}
}
\maketitle              
\begin{abstract}
 We consider a new variant of the multi-robot task allocation problem - Inverse Risk-sensitive Multi-Robot Task
Allocation (IR-MRTA). 
 "Forward" MRTA - the process of deciding which robot should perform a task given the reward (cost)-related parameters, is widely studied in the multi-robot literature. In this setting, the reward (cost)-related parameters are assumed to be already known: parameters are first fixed offline by domain experts, followed by coordinating robots online. What if we need these parameters to be adjusted by non-expert human supervisors who oversee the robots during tasks to adapt to new situations? We are interested in the case where the human supervisor's perception of the allocation risk may change and suggest different allocations for robots compared to that from the MRTA algorithm. In such cases, the robots need to change the parameters of the allocation problem based on evolving human preferences. We study such problems through the lens of inverse task allocation, i.e., the process of finding parameters given solutions to the problem. Specifically, we propose a new formulation IR-MRTA, in which we aim to find a new set of parameters of the human behavioral risk model that minimally deviates from the current MRTA parameters and can make a greedy task allocation algorithm allocate robot resources in line with those suggested by humans. We show that even in the simple case such a problem is a non-convex optimization problem. We propose a Branch $\&$ Bound algorithm (BB-IR-MRTA) to solve such problems. In numerical simulations of a case study on multi-robot target capture, we demonstrate how to use BB-IR-MRTA and we show that the proposed algorithm achieves significant advantages in running time and peak memory usage compared to a brute-force baseline.

\end{abstract}
\section{Introduction}
Multi-robot task Allocation (MRTA) is a fundamental problem in multi-robot systems \cite{korsah2013comprehensive,khamis2015multi,chakraa2023optimization}. It determines which robots execute particular tasks to collaboratively achieve a system-level goal \cite{korsah2013comprehensive}. It can generally be cast as a combinatorial optimization problem with constraints and is usually NP-hard. 

In the classic MRTA pipeline, we first design reward/cost functions for allocations using domain knowledge. This phase is typically offline. Then, we deploy the robots for tasks by solving a combinatorial optimization problem. A missing piece in such a pipeline is how to incorporate feedback from human supervisors into the task allocation process when we deploy the robots in the field, and human supervisors stay in the loop to watch over the team. Considering human feedback in multi-robot task allocation is crucial especially when there is uncertainty. First, humans can provide valuable insights and contextual knowledge that may not be fully captured by pre-programmed algorithms, especially in complex or unforeseen situations. Second, involving humans in the decision-making process helps to align robot operations with human preferences (and potentially ethical standards). In the risk-sensitive multi-robot task allocation problem, human supervisors may evaluate the risk associated with a task allocation schema in a way that is different from what statistical models may assess due to either humans having better situation understanding or their personal preference. As a result,    
humans may suggest different (possibly better) allocations of tasks compared to those obtained from solving pre-designed optimization problems. In such cases, it is undesirable to stop the team and redesign the optimization problem. Moreover, the human supervisors in the loop may not be optimization experts, and they may not know how to modify the problems to distill their new insights. Instead, we want the robot team to have the capability to adjust the parameters of the problem in a \textit{minimal} way to accommodate human suggestions. The reason for the minimal change is that the current parameters of the problem are already the embodiment of much expertise and historical data, and we should not give them away in favor of a few new suggestions.  We consider such problems from the perspective of the inverse problem of risk-sensitive task allocation.

Specifically, for a multi-robot task allocation problem with parameterized risk constraints/objectives \cite{asghar2023risk,nam2016analyzing,yang2017algorithm}, if the parameters are known, we can use an approximation algorithm to solve the problem. We call such a process Forward Task Allocation (FTA). By contrast, if we have a solution obtained using approximation algorithms, we want to find the corresponding parameters related to risk such that when we use those parameters to solve FTA problems the resulting solutions match the known solutions. We call such a process Inverse Risk-sensitive Multi-Robot Task Allocation (IR-MRTA). Such problems fall into the more general class of inverse optimization problems, which are widely used to decode human decision-making processes \cite{terekhov2010analytical,tsirakos1997inverse,sandholtz2023inverse} for robotic applications. 
By formulating and solving IR-MRTA we can incorporate human suggestions into the task allocation process. 
To our knowledge, this is the first paper that introduces such a risk-sensitive inverse problem formulation to MRTA. Further, we introduce a Branch and Bound (BB) algorithm to solve IR-MRTA.

The main contributions of this paper are:
\begin{itemize}
    \item We formulate a new variant of MRTA named IR-MRTA and propose a novel algorithm to solve IR-MRTA.
    \item We show how to use IR-MRTA in multi-robot task allocation to account for human suggestions.
    \item The proposed formulation and algorithm are validated extensively through simulation. 
\end{itemize}

\section{Related Work}

\subsubsection{Multi-Robot Task Allocation} 
MRTA problems have been widely studied in the context of multi-robot systems. Most existing research focuses on how to develop efficient (scalable to a large number of robots) and effective (optimal or close to optimal) strategies, which can be centralized or decentralized, to solve deterministic allocation problems. Uncertainty in MRTA is also studied using either risk-neutral approaches \cite{prorok2019redundant,choudhury2022dynamic} or risk-averse approaches \cite{nam2016analyzing,rudolph2021desperate,yang2017algorithm}. The differences between our work and existing work lie in two aspects. First, most existing work considers the forward problem, i.e., given the specified parameters in the objective and the constraints, find a solution. By contrast, our focus is the inverse problem, i.e., given a solution, find parameters in the constraints. Second, our formulation of MRTA incorporates a parameterized human behavioral model, which describes how humans perceive uncertainty. 
\vspace*{-3.8ex}
\subsubsection{Inverse Optimization}
Inverse Optimization (IO) is widely used to learn human behavior,  for example, using human walking gaits for humanoid locomotion and using human grasping skills for robot arm manipulation. In the related literature, IO is usually referred to as Inverse Optimal Control (IOC) or Inverse Reinforcement Learning (IRL).  Existing frameworks for IOC/IRL are usually designed for one robot and their multi-robot counterparts are still in their infancy \cite{vsovsic2016inverse,yu2019multi,wang2018competitive}. The main difference between our work and the existing IOC/IRL is that our forward problem is combinatorial with discrete decision variables rather than continuous optimization. Such combinatorial cases in inverse optimization are less well studied. Since combinatorial optimization is widely used for multi-robot coordination, e.g., task allocation and vehicle routing and scheduling, the research for such Inverse Combinatorial Optimization (ICO) \cite{heuberger2004inverse} will provide more tools for multi-robot systems. 
In ICO, Inverse Integer Optimization (IIO) is widely studied \cite{schaefer2009inverse}. However, IIO can not deal with the inverse problem studied in this paper as shown in Sec. \ref{sec:prb_frml} since it involves a nonlinear and non-convex constraint.
\vspace*{-3.8ex}
\subsubsection{Human Multi-Robot Interaction}
Our research is closely related to two directions.  One is human-in-the-loop multi-robot coordination \cite{wang2018trust,diaz2015distributed,xu2023leveraging,lin2015experiments}. Our work is different from these works in two aspects: first, we consider the problem for a novel task allocation problem; second, we develop a new paradigm by using ICO to incorporate human suggestions. Another closely related direction is human preference learning \cite{wilde2020improving,biyik2019asking,wilde2020active}, in which humans are typically presented with two solutions iteratively, requiring them to compare these options during each iteration. The majority of research in this area concentrates on developing efficient methods for generating queries to facilitate the learning of human preference parameters. By contrast, our ICO does not involve an iterative process. Humans suggest an allocation that they think the \textit{approximation algorithm} should output. We develop a framework that minimally modifies the parameters of the problem to match the human suggestions.

\section{Preliminaries}
We begin with the notations and conventions used in this paper and introduce some background before formally defining the inverse problem.   
We use calligraphic fonts to denote sets (e.g. $\mathcal{A}$). We denote $\norm{\cdot}$ as the $\ell^2$-norm. A set is ordered if it preserves the order in which elements are inserted. For an ordered set $\mathcal{S}$, we use $\mathcal{S}[i]$ and $\mathcal{S}[1:i]$ to refer to its $i$-th element and its first $i$ elements, respectively. For an unordered set, we call each permutation of the set an \textit{ordering} of the set.

\subsection{Forward Problem: Risk-sensitive Multi-Robot Task Allocation}
We first introduce a risk-sensitive multi-robot task allocation formulation \cite{asghar2023risk}. In this formulation, each allocation is associated with a reward and a probability of staying intact after executing the allocated task. The objective is to maximize the sum of rewards and guarantee the likelihood of all the allocated robots being intact being greater than a threshold. Then, we will introduce the behavioral model of humans (i.e., how humans perceive uncertainty) into the formulation, leading to the behavioral risk-sensitive task allocation. 

{Risk-sensitive task allocation} is formulated as 
\begin{subequations}
  \begin{align}
    \max_{\bm{x}}~ & \sum_{i,j} r_{i, j} x_{i, j} \\
    \rm{s.t.} \quad & \sum_{i}^{n_r} x_{i, j} \leq 1, \forall j, \quad \sum_{j}^{n_t} x_{i, j} \leq 1, \forall i
    \label{eq:task_allocation_risk:constraint1}\\
    & \prod_{i, j} ((1-x_{i, j})+x_{i,j}p_{i, j}) \geq \delta \label{eq:task_allocation_risk:constraint2}\\
    & \sum_{i, j} x_{i, j} \leq n_r \label{eq:task_allocation_risk:constraint3}\\
    & x_{i,j} \in \{0, 1\}\label{eq:task_allocation_risk:constraint4},
  \end{align}
\end{subequations}
where $r_{ij}$ denotes the reward of allocating robot $i$ to task $j$; the constraint \eqref{eq:task_allocation_risk:constraint1} enforces that one robot can be allocated to up to one task and one task can be allocated at most one robot; in the constraint \eqref{eq:task_allocation_risk:constraint2}, if the robot $i$ is allocated to the task $j$ (i.e., $x_{ij}=1$), $((1-x_{i, j})+x_{i,j}p_{i, j})$ is equal to $p_{ij}$, otherwise is 1. The product is the likelihood of all the allocated robots being intact and $\delta$ is a threshold; Eq. \eqref{eq:task_allocation_risk:constraint3} is a constraint on the number of available robots; Eq. \eqref{eq:task_allocation_risk:constraint4} specifies that $x_{ij}$ is a binary variable.

\textbf{Human Perceived Uncertainty} According to research in behavioral economics, humans usually perceive uncertainty in an irrational fashion, especially in evaluating outcomes \cite{prelec1998probability,bach2012knowing}. We introduce such a concept into the risk-sensitive task allocation to account for the observation that if we need to solve a sequence of task allocation problems to coordinate robots, humans may change their views on risks based on the results of robots after executing the allocated tasks. Humans may become more conservative or more aggressive. Specifically, we consider the probability weight function $w:[0, 1] \to [0, 1]$ which maps a probability to another and can be used to model human irrationality on uncertainty \cite{prelec1998probability,bach2012knowing}. In this paper, we use the popular Prelec's model \cite{prelec1998probability} for its simplicity
\begin{equation}
    w(p) = e^{-\beta(-\log p)^{\alpha}}, \alpha > 0, \beta > 0, w(0) = 0.
\end{equation}
More details on the intuition of such a model are given in Sec. \ref{sec:qualitative}.

Using Prelec's model, the constraint in \eqref{eq:task_allocation_risk:constraint2} becomes
\begin{equation}
    \prod_{i, j} ((1-x_{i, j})+x_{i,j}{w(p_{i, j})}) \geq \delta \label{eq:behavioral:constraint2}
\end{equation}

By taking logarithmic operation on both sides, Eq. \eqref{eq:behavioral:constraint2} can be equivalently represented as 
\begin{equation}
    -\sum \log(w(p_{i, j})) x_{i,j} \leq -\log \delta,
\end{equation}
i.e., 
\begin{equation}
    \sum \beta(-\log p_{i,j})^{\alpha} x_{i,j} \leq ~-\log \delta.
\end{equation}

The forward task allocation problem can be summarized as follows.
\begin{subequations}
    \begin{align}
        \max_{\bm{x}}~ & \sum_{i,j} r_{i, j} x_{i, j} \label{eq:task_allocation_objective}\\
        \rm{s.t.} \quad & \sum_{i}^{n_r} x_{i, j} \leq 1, \forall j, \quad \sum_{j}^{n_t} x_{i, j} \leq 1, \forall i \label{eq:behavioral_risk1:constraint1}\\
         & \sum_{i, j} \beta(-\log p_{i,j})^{\alpha} x_{i,j} \leq ~-\log \delta \label{eq:behavioral1:constraint2}\\
        & \sum_{i, j} x_{i, j} \leq n \\
        & x_{i,j} \in \{0, 1\}
    \end{align}
\end{subequations}

This can be viewed as a standard task allocation with one extra knapsack constraint (Eq. \eqref{eq:behavioral1:constraint2}), rendering the problem NP-hard \cite{ozbakir2010bees,salkin1975knapsack}. In this paper, we assume that we will use 
a simple greedy algorithm to solve the problem as shown in the Algorithm \ref{algorithm:greedy}. At each step, we will select one allocation based on a score which is the allocation reward ($r_{ij}$) divided by its associated cost ($\beta (-\log p_{i,j})^{\alpha}$).

\begin{algorithm}[ht]\label{algorithm:greedy}
    \caption{Greedy Algorithm}
    \SetKwInOut{Input}{Input}
    \SetKwInOut{Output}{Output}
    \Input{A TA-BRC instance 
    }
    \Output{
    A task allocation plan 
    }
    $\mathcal{S} \gets \emptyset$ \\
    \While{$\abs{\mathcal{S}} < n$}{
    for unallocated robots, find the allocation with the largest cost-scaled reward 
    $(i^*, j^*) = \argmax_{i, j}~\frac{r_{i,j}}{\beta(-\log p_{i,j})^{\alpha}}$ \\
    
    \eIf{$\sum_{(i, j)\in \mathcal{S} \cup \{(i^*, j^*)\}} \beta(-\log p_{i,j})^{\alpha} \leq ~-\log \delta $ }{
     $\mathcal{S} \gets \mathcal{S} \cup \{(i^*, j^*)\}$
    }
    {
    break
    }
    }
    return $\mathcal{S}$
\end{algorithm}

\section{Problem Formulation}\label{sec:prb_frml}
\begin{problem}[Inverse Risk-sensitive Multi-Robot Task Allocation (IR-MRTA)]\label{problem:IRMRTA}

Given a suggested feasible allocation result $\hat{\bm{x}} \in \{0, 1\}^{n_r \times n_t}$ from human, find new parameters $\hat{\alpha}, \hat{\beta}, \hat{\delta}$ for the human behavior model such that the weighted sum of the distances between the original parameters ($\alpha, \beta, \delta$) and the new parameters is minimized, and the suggested allocation $\hat{\bm{x}}$ becomes the output of the greedy algorithm using $(\hat{\alpha}, \hat{\beta}, \hat{\delta})$. Mathematically, such an inverse task allocation problem can be formulated as follows: 
      \begin{subequations}
          \begin{align}
            \min_{\hat{\alpha}, \hat{\beta}, \hat{\delta}}~ & w_{\alpha}\norm{ \hat{\alpha}-\alpha} + w_{\beta}\norm{ \hat{\beta}-\beta}+w_{\delta}\norm{ \hat{\delta}-\delta}\label{eq:main_inverse_obj}\\
            \rm{s.t.} \quad & \bm{x}^*(\hat{\alpha}, \hat{\beta}, \hat{\delta}) = \argmaxgreedy_{\bm{x}} f(\bm{x} \mid \hat{\alpha}, \hat{\beta}, \hat{\delta})  \label{eq:inverse_risk:solution}\\ 
            & \hat{\bm{x}} = \bm{x}^*(\hat{\alpha}, \hat{\beta}, \hat{\delta})\label{eq:inverse_risk:constraint1}\\
            & \hat{\alpha} \in [\hat{\alpha}_{min}, \hat{\alpha}_{max}],   \hat{\beta} \in [\hat{\beta}_{min}, \hat{\beta}_{max}], \hat{\delta} \in [\hat{\delta}_{min}, \hat{\delta}_{max}] ,
            \end{align}
    \end{subequations}
    where $f(\bm{x} \mid \hat{\alpha}, \hat{\beta}, \hat{\delta})$ is the objective in Eq. \eqref{eq:task_allocation_objective} and $\hat{\alpha}, \hat{\beta}, \hat{\delta}$ is incorporated to highlight that the optimization will rely on these parameters; $\bm{x}^*(\hat{\alpha}, \hat{\beta}, \hat{\delta})$ in Eq. \eqref{eq:inverse_risk:solution} is the solution returned by the greedy algorithm (Algorithm \ref{algorithm:greedy}) using parameters $\hat{\alpha}, \hat{\beta}, \hat{\delta}$ and Eq. \eqref{eq:inverse_risk:constraint1} enforces the solution to be equal to the human suggested allocation; $w_{\alpha}, w_{\beta}$ and $w_{\delta}$ are three hyperparameters.  
\end{problem}

\subsection{Motivating Case Study: Multi-Robot Target Capture}\label{sec:motivating_case_study}
There are $n_r$ robots in the task area $\mathcal{E}$. Robots are represented as discs with different sizes. The larger the disc, the more energy it will consume to finish the same task. A team of $n_t$ non-strategic targets with $n_t \leq n_r$ will occasionally enter the $\mathcal{E}$. The targets are also represented as discs with different sizes. These targets are of value to us, and we want to capture them by robots. We consider the case where we must allocate one robot to each target for successful capture. When the robot captures a target, the target can cause damage to the robot with a probability that depends on the relative size of each other. When the robot is larger compared to the target, the likelihood of being damaged is small. by contrast, when the robot is relatively small compared to the target, the damage probability will be higher. After going through the damage, the robot is still functional but needs inspection and repair. The high-level goal here is to allocate robots to capture more targets to maximize the reward and try to avoid damage. 

Specifically, the reward of allocating robot $i$ to target $j$ is defined as
\begin{equation}
    r_{ij} = r_j - \xi_c \hat{c}_{ij} - \xi_d (1-p_{ij}) d_i,
\end{equation}
where $r_j$ is the reward term for capturing the target $j$; $\xi_c$ is a scaling factor for the motion cost; $\hat{c}_{ij}$ is the estimated motion cost (e.g., energy consumption).  Suppose the target is non-strategic and moves with constant velocities, the motion cost will be proportional to the travel distance to intercept the target; $\xi_d$ is a scaling factor for repair cost after damage; $p_{ij}$ is the probability of without being damaged; and $d_i$ is the repair cost for robot $i$; $r_{ij}$ is designed to be positive.

$p_{ij}$ should capture that the relative size between the robot and the target will affect the probability of being damaged. We use the following function in this paper
\begin{equation}\label{eq:damage_prob}
    p_{ij} = \frac{1}{1+e^{k(s_j-s_i)}},
\end{equation}
where $k>0$ is a parameter; $s_i$ and $s_j$ are sizes for robot $i$ and target $j$ respectively.

After a few rounds of capture, if no robots get damaged, humans may think the Eq. \eqref{eq:damage_prob} underestimates the probability of being intact and suggest a more aggressive allocation (e.g., replace a larger robot with a smaller robot to reduce motion cost). We need to solve the inverse problem to determine the parameters that can lead to such suggested decisions.

\section{Algorithm}
Let us first consider the case where the human suggestion is ordered, i.e., the suggested allocation is given sequentially.  
Let ${\mathcal{S}} = \{{(i_1, j_1)}, {(i_2, j_2)}, \ldots,  {(i_m, j_m)}\}~(m \leq n_r)$ be the ordered set of allocations corresponding to the human suggestions. We will denote $\mathcal{S}^i$ and $\mathcal{S}^j$ as the ordered sets $\{{i_1}, {i_2}, \ldots,  {i_m}\}$ and $\{{j_1}, {j_2}, \ldots,  {j_m}\}$ respectively. Recall that when we use the greedy algorithm (Algorithm \ref{algorithm:greedy}) to determine allocations, we will find the allocation with maximum score each step and make the new allocation will satisfy the knapsack constraint. Suppose that the human suggests that $\mathcal{S}$ should be the output of using the greedy algorithm, 
such an ordered set is equivalent to a set of inequalities as follows. 
\begin{subequations}
    \begin{align}
       & \frac{r_{i_1,j_1}}{\hat{\beta}(-\log p_{i_1,j_1})^{\hat{\alpha}}} \geq  \frac{r_{i, j}}{\hat{\beta}(-\log p_{i,j})^{\hat{\alpha}}}, \forall i, j \\
       & \hat{\beta}(-\log p_{i_1,j_1})^{\hat{\alpha}} \leq -\log \hat{\delta}
    \end{align}

    \begin{align}
         &\frac{r_{i_2,j_2}}{\hat{\beta}(-\log p_{i_2, j_2})^{\hat{\alpha}}} \geq  \frac{r_{i, j}}{\hat{\beta}(-\log p_{i,j})^{\hat{\alpha}}}, \forall i(j) \not\in \mathcal{S}^{i(j)}[1:1]\\
        &\sum_{k=1}^2 \hat{\beta}(-\log p_{i_k,j_k})^{\hat{\alpha}} \leq -\log \delta
    \end{align}
    $$\cdots$$
     \begin{align}
         &\frac{r_{i_m,j_m}}{\hat{\beta}(-\log p_{i_m, j_m})^{\hat{\alpha}}} \geq  \frac{r_{i, j}}{\hat{\beta}(-\log p_{i,j})^{\hat{\alpha}}}, \forall i(j) \not\in \mathcal{S}^{i(j)}[1:m-1] \\
        &\sum_{k=1}^m \hat{\beta}(-\log p_{i_k,j_k})^{\hat{\alpha}} \leq -\log \hat{\delta} \label{eq:risk_constraints_inverse}
    \end{align}
\end{subequations}

These inequalities correspond to the greedy selection process as described in Algorithm \ref{algorithm:greedy} and can be simplified as follows. First, we can observe that all the inequalities corresponding to the knapsack constraint in Eq. \eqref{eq:behavioral1:constraint2} can be summarized as Eq. \eqref{eq:risk_constraints_inverse} since other inequalities are included in this inequality. Second, we can transform the inequalities corresponding to the score comparison and greedy selection into linear constraints by taking log operations on both sides of the inequalities. Therefore, the inequalities above can be simplified as follows. 
\begin{subequations}
    \begin{align}\label{eq:linear1}
        \log r_{i_1,j_1} + \hat{\alpha} \cdot \log (\log \frac{1}{p_{i, j}}) \geq \log  r_{i, j} + \hat{\alpha} \cdot \log (\log \frac{1}{p_{i_1, j_1}})
    \end{align}
    \begin{equation}\label{eq:linear2}
          \begin{aligned}
        \log r_{i_2,j_2} + \hat{\alpha} \cdot \log (\log \frac{1}{p_{i, j}}) \geq \log  r_{i, j} +  \hat{\alpha} \cdot \log (\log \frac{1}{p_{i_2, j_2}}), \\
        \forall i(j) \not\in \mathcal{S}^{i(j)}[1:1]
    \end{aligned}
    \end{equation}
    $$\cdots$$
    \begin{equation}\label{eq:linear3}
         \begin{aligned}
        \log r_{i_m,j_m} + \hat{\alpha} \cdot \log (\log \frac{1}{p_{i, j}}) \geq \log  r_{i, j} +  \hat{\alpha} \cdot \log (\log \frac{1}{p_{i_m, j_m}}), \\
       \forall i(j) \not\in \mathcal{S}^{i(j)}[1:m-1]
        \end{aligned}
    \end{equation}
    \begin{equation}
        \sum_{k=1}^m \hat{\beta}(-\log p_{i_k,j_k})^{\hat{\alpha}} \leq -\log \hat{\delta}
    \end{equation}
\end{subequations}

In sum, the ordered set case of IR-MRTA can be summarized as the following problem. 
\begin{problem}[Ordered-IR-MRTA (O-IR-MRTA)]\label{problem:O-IRMRTA}
    \begin{subequations}
  \begin{align}
    \min_{\hat{\alpha}, \hat{\beta}, \hat{\delta}}~ & w_{\alpha}\norm{ \hat{\alpha}-\alpha} + w_{\beta}\norm{ \hat{\beta}-\beta }+w_{\delta}\norm{ \hat{\delta}-\delta}\\
    \rm{s.t.} \quad & \hat{\alpha} \in \mathcal{A}, \hat{\alpha} \in [\hat{\alpha}_{min}, \hat{\alpha}_{max}],   \hat{\beta} \in [\hat{\beta}_{min}, \hat{\beta}_{max}], \hat{\delta} \in [\hat{\delta}_{min}, \hat{\delta}_{max}]  \\
    & \sum_{k=1}^m \hat{\beta} (-\log p_{i_k,j_k})^{\hat{\alpha}} \leq -\log \hat{\delta}, \label{eq:non-convex-constraint}
    \end{align}
\end{subequations}
where $\mathcal{A}$ is a convex set corresponding constraints described from Eq. \eqref{eq:linear1} to Eq. \eqref{eq:linear3}.
\end{problem}

It should be noted that Eq. \eqref{eq:non-convex-constraint} is not a convex constraint. In the following, we will show how to solve such a non-convex optimization problem using the branch and bound paradigm. The key idea is to partition the feasible set into convex sets and find the lower/upper bound for each, which is then used to decide the expansion of the search tree. The algorithm will refine the partition until the search depth reaches the termination condition.  

Specifically, we will partition $[\hat{\beta}_{min}, \hat{\beta}_{max}]$ and $[\hat{\delta}_{min}, \hat{\delta}_{max}]$ into smaller convex subsets $\mathcal{B}=[\ubar{\beta}, \bar{\beta}]$ and $\mathcal{D}=[\ubar{\delta}, \bar{\delta}]$. 
 We aim to solve the following problem.
\begin{subequations}
  \begin{align}
    \min_{\hat{\alpha}, \hat{\beta}, \hat{\delta}}~ & w_{\alpha}\norm{\hat{\alpha}-\alpha } + w_{\beta}\norm{\hat{\beta}-\beta}+w_{\delta}\norm{\hat{\delta}-\delta}\\
    \rm{s.t.} \quad & \hat{\alpha} \in \mathcal{A}, \quad \hat{\beta} \in \mathcal{B},  \quad \hat{\delta} \in \mathcal{D} \\
    & \sum_{k=1}^m (-\log p_{i_k,j_k})^{\hat{\alpha}} \leq -\frac{1}{\hat{\beta}}\log \hat{\delta} \label{eq:knapsack_original}
    \end{align}
\end{subequations}
It should be noted that $-\frac{1}{\hat{\beta}}\log \hat{\delta} $ is a monotone decreasing function w.r.t. both $\hat{\beta}$ and $\hat{\delta}$, i.e., $-\frac{1}{\hat{\beta}}\log \hat{\delta} \leq -\frac{1}{\ubar{\beta}}\log \ubar{\delta}$. Using this observation, we can get the convex relaxation to compute the lower bound of the optimal objective value by replacing Eq. \eqref{eq:knapsack_original} with Eq. \eqref{eq:knapsack_relaxed}.  

\begin{equation}\label{eq:knapsack_relaxed}
    \sum_{k=1}^m (-\log p_{i_k,j_k})^{\alpha} \leq -\frac{1}{\ubar{\beta}}\log \ubar{\delta} 
\end{equation}
In such convex relaxation, the constraint is imposed only on $\hat{\alpha}$. If it is not feasible, it suggests that the original problem is not feasible. If it is feasible, we can find a feasible solution by setting $\beta = \ubar{\beta}, \delta=\ubar{\delta}$ and solve the following problem.

\begin{subequations}
  \begin{align}
    \min_{\hat{\alpha}}~ & w_{\alpha}\norm{\hat{\alpha}-\alpha} + w_{\beta}\norm{\ubar{\beta} - \hat{\beta}}+w_{\delta}\norm{\ubar{\delta} - \hat{\delta}}\\
    \rm{s.t.} \quad & \hat{\alpha} \in \mathcal{A}  \\
    & \sum_{k=1}^m (-\log p_{i_k,j_k})^{\hat{\alpha}} \leq -\frac{1}{\bar{\beta}}\log \ubar{\delta} 
    \end{align}
\end{subequations}

\begin{algorithm}[ht!]\label{algorithm:BB_nonconvex}
    \caption{B\&B for O-IR-MRTA (BB-O-IR-MRTA)}
    \SetKwInOut{Input}{Input}
    \SetKwInOut{Output}{Output}
    \SetKwProg{Fn}{Function}{:}{}
    \Input{An O-IR-MRTA instance
    }
    \Output{
    An approximate global solution ${\alpha}^*, {\beta}^*, {\delta}^*$
    }
    Tree $\gets$ empty tree \\
    Tree.add\_node(node\_ID=0, LB=$-1$, UB=$\infty$)\\
    Tree.sol $\gets$ null \# best solution found \\
    Tree.obj\_ub $\gets \infty$,  Tree.obj\_lb  $\gets 0$ \\
    $Q \gets$  queue \\
    $Q$.insert(node\_ID=0)\\
    \While{$Q$ is not empty or Tree.depth reaches set value}{
    $v_p \gets Q$.pop()  \\
    $\mathcal{V}_c \gets$ generate\_child\_nodes($v_p$) \\
    \For{each child node $v_c \in \mathcal{V}_c$}{
    LB\_feasibility, LB $\gets$ lower\_bound($v_c$)\\
    UB\_feasibility, UB $\gets$ upper\_bound($v_c$)\\
    \eIf{not LB\_feasibility}{
     continue 
    }
    {
      \If{LB $>$ Tree.obj\_ub}
      {continue}
      \If{UB $<$ Tree.obj\_ub}
      {
      Tree.sol $\gets$ solution at $v_c$ \\
      Tree.obj\_ub $\gets$ UB at $v_c$ \\
      
      } 

      Q.insert($v_c$) \\
      Tree.add\_node($v_c$)
    }
    }
    }
    
    return Tree.sol
\end{algorithm}

\begin{figure}[t]
    \centering
    \subfloat[]{
    \includegraphics[width=0.43 \textwidth]{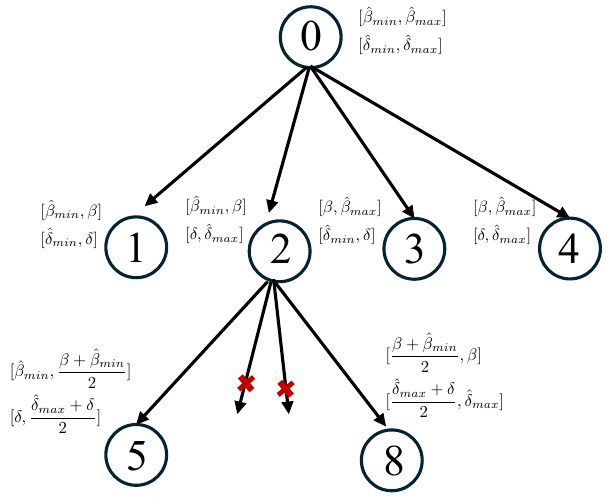}
    \label{fig:BB_O_IRMRTA}
    } 
    \subfloat[]{
    \includegraphics[width=0.43\textwidth]{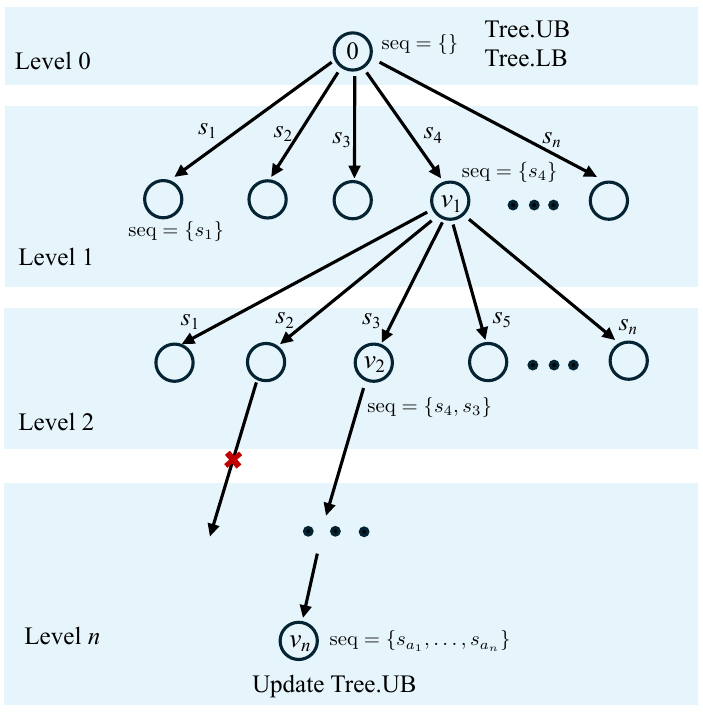}
    \label{fig:BB}
    }
    \caption{
     (a) One search iteration of the proposed subroutine algorithm BB-O-IR-MRTA. (b) One iteration of the proposed BB-IR-MRTA algorithm. }
    \label{fig:algorithm_explanation}
\end{figure}

Details are given in Algorithm \ref{algorithm:BB_nonconvex}. 
In lines 1-6, we initialize the search process. Tree.sol is used to track the best solution found so far and Tree.obj\_ub is the corresponding upper bound. In the while loop, we first extract a node from the queue (line 8). Then, we will generate some child nodes by dividing the sets of $\beta$ and $\delta$ (line 10). We choose the following splitting strategies to create child nodes. As shown in Fig. \ref{fig:BB_O_IRMRTA}, at the root node, we divide the set by using the original parameter ($\beta$ and $\delta$) as a separator and we will generate four child nodes. For the rest, we generate four child nodes by evenly dividing each interval into two parts (from node 2 to nodes 5 and 8). After this step, for each generated child node, we will find its lower bound and upper bound (lines 11-12) considering the relaxation described above. If it is infeasible to find its lower bound (lines 13-15), it suggests that the problem is infeasible for this node and we can prune this branch away (red cross in Fig. \ref{fig:BB_O_IRMRTA}). Otherwise, we check whether the lower bound is greater than the best upper bound identified so far (lines 16-18). If it is, it suggests that we do not need to further branch on this node since it will not result in a solution better than the one identified so far. If the upper bound is less than the global upper bound which suggests that we have found a better solution, we will update the global solution and the upper bound (lines 19-22). This node will be inserted into the queue for further branching (lines 23-24). After updating all the child nodes, we will check whether the termination condition is reached to decide whether we should further branch the search tree (line 7).  Upon termination, the solution will satisfy the property outlined in Theorem  \ref{theorem:BB_nonconvex}.

\begin{theorem}\label{theorem:BB_nonconvex}
    Given a feasible problem instance as described in Problem \ref{problem:O-IRMRTA}, Algorithm \ref{algorithm:BB_nonconvex} returns a solution, $sol$, satisfying 
    \begin{equation*}
        {g(sol)-g^*} \leq  w_{\beta} \max(\frac{\beta-\hat{\beta}_{min}}{2^{n_d-1}}, \frac{\hat{\beta}_{max}-\beta}{2^{n_d-1}}) 
               + w_{\delta} \max(\frac{\delta-\hat{\delta}_{min}}{2^{n_d-1}}, \frac{\hat{\delta}_{max}-\delta}{2^{n_d-1}}), 
    \end{equation*}
    where $g(\cdot)$ is the objective in Eq. \eqref{eq:main_inverse_obj} and $g^*$ is the optimal objective to Problem \ref{problem:O-IRMRTA}; $w_{\beta}, w_{\delta}, \beta, \delta, \hat{\beta}_{min}, \hat{\beta}_{max}, \hat{\delta}_{min},  \hat{\delta}_{max}$ are defined in the Problem \ref{problem:O-IRMRTA}.
\end{theorem}

The proof of Theorem \ref{theorem:BB_nonconvex} and the analysis of how the optimality gap depends on the depth of the search tree are given in the Appendix. We provide a high-level overview of the proof here. 
In Algorithm \ref{algorithm:BB_nonconvex}, we grow the search tree in a breadth-first search fashion: we will search all the possible combinations in a certain depth and then move to the next depth. At each depth level, the leaf nodes contain all the possible combinations of subsets of $\hat{\beta}$ and $\hat{\delta}$. Suppose that the best solution identified belongs to a node $v_u$, i.e., $v_u.UB = g(sol)$ and $v_u.UB \geq v.UB$ for all leaf nodes.    
Suppose that the optimal solution will be in a node $v_{OPT}$  which is not known to us. But what we know is that $v_{OPT}.LB \leq g^* \leq v_{OPT}.UB$. The proof is on how to relate the gap $g(sol)-g^*$ to the set sizes of each leaf node.

 \begin{algorithm}[ht!]\label{algorithm:BB_algorithm}
    \caption{B\&B for IR-MRTA (BB-IR-MRTA)}
    \SetKwInOut{Input}{Input}
    \SetKwInOut{Output}{Output}
    \SetKwProg{Fn}{Function}{:}{}
    \Input{human suggestion $\hat{\mathcal{S}}$, tolerance $\epsilon$
    }
    \Output{$\hat{\bm{\theta}}=[\hat{\alpha}, \hat{\beta}, \hat{\delta}]$}
    Tree $\gets$ empty tree \quad \# Initialize a search tree \\
    Tree.UB $\gets$ a large number \\
    Tree.add\_node(node\_ID = 0, sequence = $\{\}$) \\
    \nonl \# Initialize an empty stack for Depth First Search \\
    Stack $\gets$ empty stack \\
    Stack.push(Tree.get\_node(node\_ID = 0)) \\
    \While{Stack is not empty}{
    $u$ $\gets$ Stack.pop() \\
    PQ $\gets$ priority\_queue() \\
    \For{$s \in \hat{\mathcal{S}} \setminus {u\rm{.seq}}$ }
    {
    feasible, obj, $\hat{\bm{\theta}}$ $\gets$ O-IR-MRTA($u\rm{.seq} + \{s\}$)\\
        \If{feasible}{
        PQ.insert($s$, $\hat{\bm{\theta}}$, priority\_value=obj)   
        }
    }
    \While{PQ is not empty}{
        $s$, $\hat{\bm{\theta}}$, obj $\gets$ PQ.pop() \\
        \If{obj $<$ Tree.UB+$\epsilon$}{
            Tree.update\_UB($u$.seq+$\{s\}$, obj) \\
            new\_node $\gets$ Tree.add\_branch($u$, $s$, obj, $\hat{\bm{\theta}}$) \\
            Stack.push(new\_node)
        } 
    }
    }
    \textbf{return} $tree$ \\
     \SetKwFunction{updateUB}{update\_UB}
    \Fn{\updateUB{$Tree, ~sequence$, obj}}{
    \If{length of $sequence$ = length of $\hat{\mathcal{S}}$}
        {
            \If{obj $<$ $Tree$.UB}
            {
                $Tree$.UB $\gets$ obj \\
                $Tree$.UB\_seq $\gets$ $sequence$
            }
        }
    }
    \textbf{end} 
    
\end{algorithm}

Next, we will consider the general case where the human suggestion is not ordered. Such a case is more practical in applications: human operators know the solutions based on their expertise and observation but they do not have the concept of the ordering of a solution set. The naive way to deal with such general cases is to enumerate all the possible orderings of the human-suggested allocation and then solve O-IR-MRTA for each. However, such an approach is not scalable w.r.t. the size of the set of the human suggestion. Next, we will introduce another BB algorithm, which will use Algorithm \ref{algorithm:BB_nonconvex} as a subroutine, to solve the general case of IR-MRTA. A similar algorithmic paradigm has been proposed in our previous work \cite{shi2024inverse} for inverse submodular maximization. The key difference is that we use a novel subroutine (Algorithm \ref{algorithm:BB_nonconvex}), which will only return an approximate optimal solution rather than the optimal one compared to \cite{shi2024inverse}. Correspondingly, we change the pruning condition (line 17) in Algorithm \ref{algorithm:BB_algorithm} and will result in a weaker result (Theorem \ref{theorem:main_BB}) compared to that in \cite{shi2024inverse}.

 The main idea is that instead of directly optimizing over the best ordering of human suggestion  $\hat{\mathcal{S}}$, we incrementally add elements to form a sequence of relaxed problems. Solving these relaxed problems can help us gradually find the upper and lower bounds of the objective of the original problem and prune the suboptimal solution. The incremental search is conducted by growing a search tree in the depth-first-search fashion. An illustrative example is shown in Fig. \ref{fig:BB}. 
 At the root node, we start with an empty sequence. For the next level (level 1) of the tree, we can add any elements in ${\hat{\mathcal{S}}}$ to the sequence to form a new node. For each such node, we will solve a problem O-IR-MRTA(\textit{seq}) using the \textit{seq} property of the node. When the \textit{seq} includes only part of elements in $\hat{\mathcal{S}}$, the objective value returned by solving O-IR-MRTA(\textit{seq}) can be viewed as the lower bound of all cases where the orderings of $\hat{\mathcal{S}}$ start with \textit{seq} since the ordering with all elements implies more constraints.

Next, we select the node (node \(v_1\) in Fig. \ref{fig:BB}) with the lowest objective value, determined by solving O-IR-MRTA(\textit{seq}), for further expansion. This objective value serves as a heuristic for guiding the search, and we employ a greedy strategy to advance to the subsequent level. The expansion process is analogous to the transition from level 0 to level 1. At node \(v_1\), the \textit{seq} contains only one element, \(s_4\). Thus, we can add any element from \(\hat{\mathcal{S}} \setminus \{s_4\}\) to form a new child node at level 2 and solve the associated O-IR-MRTA(\textit{seq}) problem. This process continues until the node's \textit{seq} encompasses all elements in \(\hat{\mathcal{S}}\), signifying that a complete ordering of \(\hat{\mathcal{S}}\) has been identified. The objective value returned from solving O-IR-MRTA(\textit{seq}) for this node will be used to update the problem's upper bound (Tree.UB in the root node), as it represents the solution for a specific ordering of \(\hat{\mathcal{S}}\). We then proceed to expand the tree by backtracking to the previous level and continuing similarly to Depth First Search (DFS). During the tree expansion, if a particular O-IR-MRTA(\textit{seq}) problem is found to be infeasible—indicating that all orderings with the prefix \textit{seq} are infeasible—or if the objective value from O-IR-MRTA(\textit{seq}) exceeds the Tree.UB+$\epsilon$, where $\epsilon$ refers to the gap in Theorem \ref{theorem:BB_nonconvex}, suggesting that all orderings with the prefix \textit{seq} will not result in an improved solution, we will prune that branch as illustrated in Fig. \ref{fig:BB} (the red cross marks the pruned branch). This pruning process accelerates the search.

The detailed procedure is outlined in Algorithm \ref{algorithm:BB_algorithm}. Initially, in lines 1-5, we set up the search tree with a root node whose \textit{seq} attribute is an empty ordered set and a stack for a DFS-style search. Within the while loop, we first pop the top element from the stack (line 7) and initialize a priority queue (line 8). Next, we iterate over each element in $\hat{\mathcal{S}}$ that is not in $u.seq$ (line 9) to check if appending this element to the existing sequence makes it feasible to solve a relaxed O-IR-MRTA problem (line 10). If infeasible, we can prune this branch, as all sequences with such a prefix are impossible. If feasible, we insert the element $s$ and the corresponding $\hat{\bm{\theta}}$ using the returned objective value as the priority value. Subsequently, we update the search tree (lines 15-22), expanding branches in increasing order of the objective value (line 16). For each candidate, we determine whether to prune it (line 17) by comparing the objective value ($obj$) with the tree’s upper bound plus a tolerance ($Tree$.UB+$\epsilon$). If $obj$ exceeds this value, we prune branches with such sequences, since any ordering of $\hat{\mathcal{S}}$ with a prefix $u.seq$ will yield an objective greater than $Tree$.UB+$\epsilon$. Given that we employ an approximation algorithm (Algorithm \ref{algorithm:BB_nonconvex}) ensuring an objective within $\epsilon$ of the optimal solution, it follows that the optimal solution for any ordering of $\hat{\mathcal{S}}$ with prefix $u.seq$ will be greater than $Tree$.UB. Hence, we prune all such branches. Conversely, if the branch should not be pruned (line 17), we update the search tree's upper bound and add the new branch to the tree (line 19). To maintain the DFS search style, we push the new nodes onto the stack (line 20). After processing all elements in the priority queue, we continue the while loop by popping the top element from the stack (line 7) and repeating the process.

\begin{theorem}\label{theorem:main_BB}
    Given a feasible problem instance as described in Problem \ref{problem:IRMRTA}, Algorithm \ref{algorithm:BB_algorithm} returns a solution, $sol$, in a finite number of iterations of the outer while loop (lines 6-23)  satisfying 
    \begin{equation*}
        {g(sol)-g^*} \leq \epsilon, 
    \end{equation*}
    where $g(\cdot)$ is the objective in Eq. \eqref{eq:main_inverse_obj} and $g^*$ is the optimal solution to Problem \ref{problem:IRMRTA}; $\epsilon$ is the gap described in Theorem \ref{theorem:BB_nonconvex}.
 \end{theorem}

Like all the algorithms that are developed within the BB paradigm, the BB-IR-MRTA algorithm is in nature enumerating all the possible combinations by incrementally adding elements. The efficiency relies on the pruning steps to remove unnecessary expansion of the search tree.
    It should be noted that the problem as described in Problem \ref{problem:IRMRTA} is an NP-hard problem in general. In the worst case, the proposed algorithms (Algorithm \ref{algorithm:BB_nonconvex} and \ref{algorithm:BB_algorithm}) will have to enumerate all the possible cases to find the solution or find that the problem is infeasible. We will experimentally evaluate the proposed algorithms in Section \ref{sec:experiment}.

\section{Experiments}\label{sec:experiment}
We validate the proposed IR-MRTA formulation and evaluate the BB-IR-MRTA algorithm in a case study on multi-robot target capture as described in Sec. \ref{sec:motivating_case_study}. We will first present a qualitative example to show how different human suggestions correspond to different parameters in the behavior models (probability weighting function). Then, we will evaluate the proposed algorithm w.r.t. its optimality, running time, and peak memory usage. 
\subsection{A Qualitative Example}\label{sec:qualitative}
\begin{figure}[t]
    \centering
    \subfloat[]{
    \includegraphics[width=0.29 \textwidth]{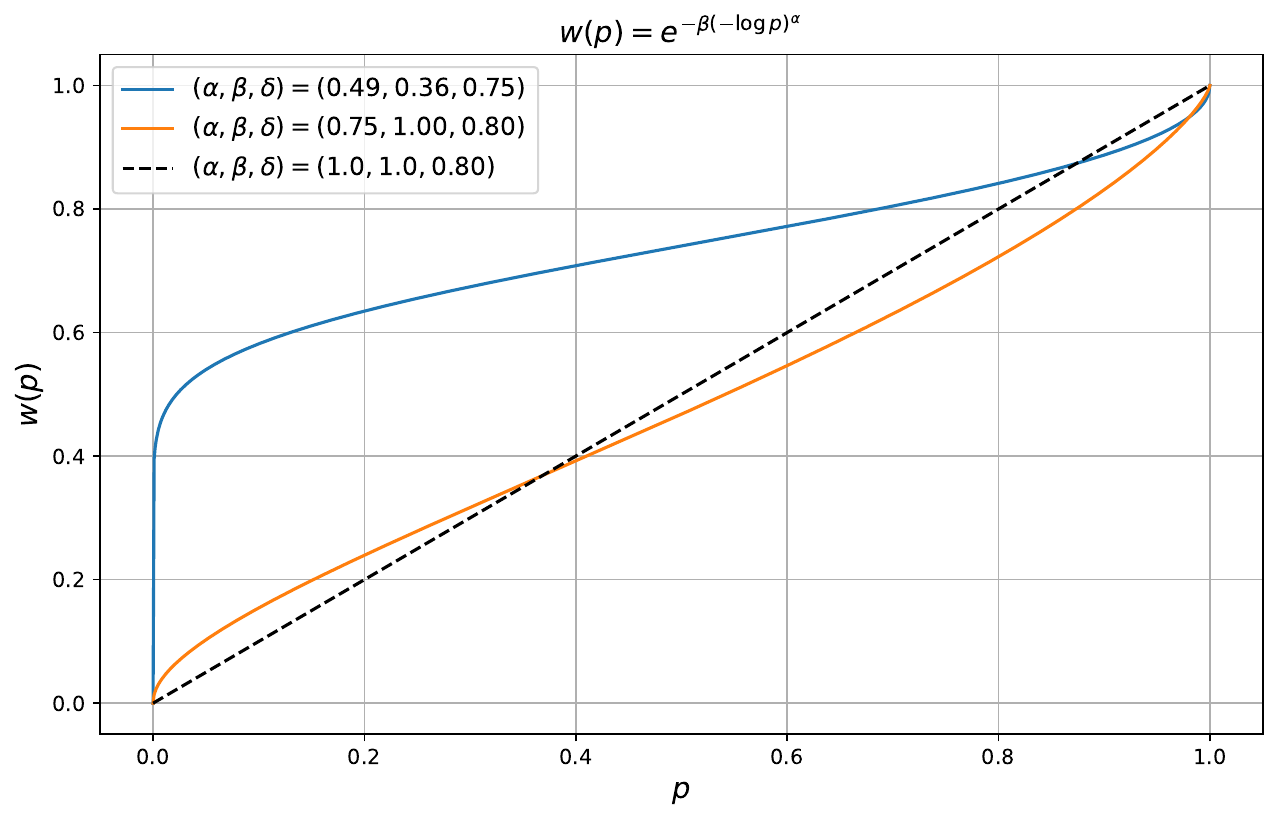}
    \label{fig:behavior_risk_model}
    } 
    \subfloat[]{
    \includegraphics[width=0.325\textwidth]{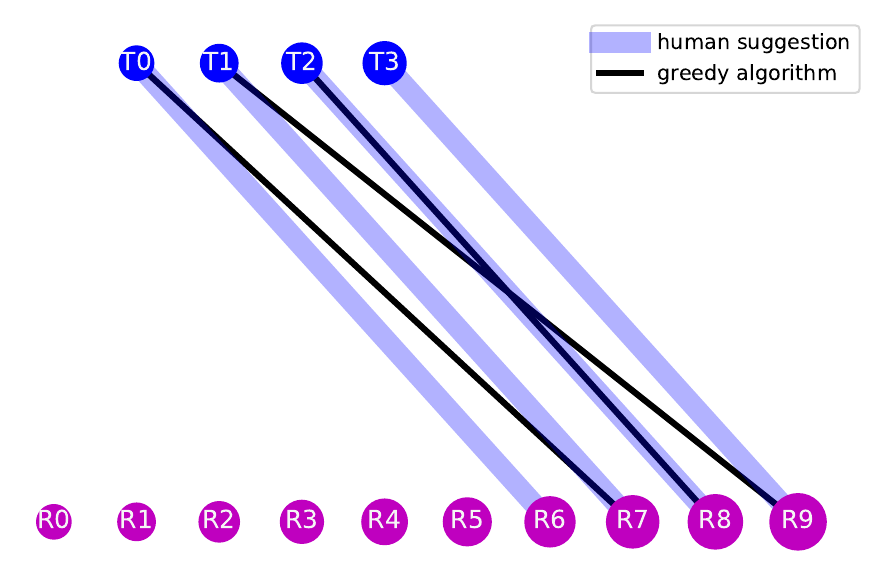}
    \label{fig:aggresive_suggestion}
    }
    \subfloat[]{
    \includegraphics[width=0.325\textwidth]{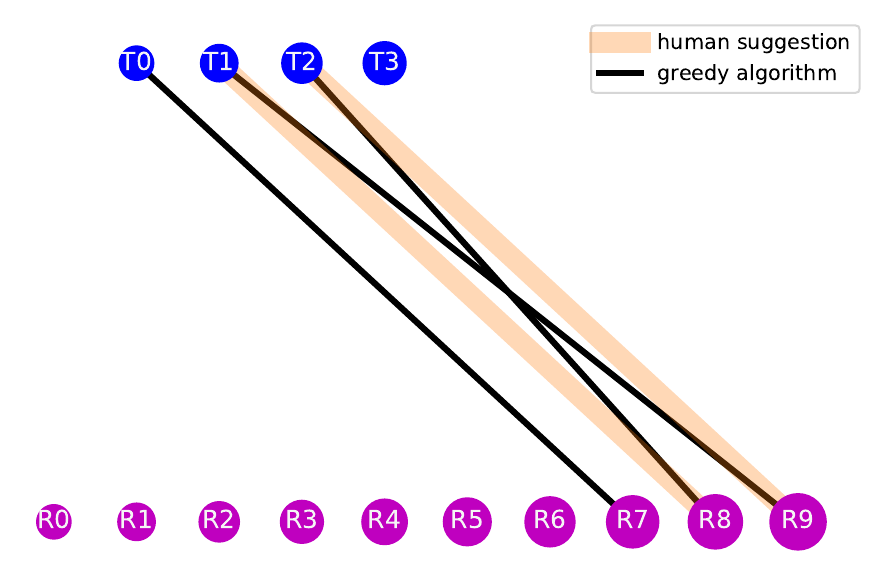}
    \label{fig:conservative_suggestion}
    }
    \caption{
     A qualitative example illustrates how to use human suggestion to identify human behavior parameters using IR-MRTA. (a) Prelec's weighting functions are identified through human suggestion. The dotted line is the case $w(p)=p$. The blue line corresponds to the suggestion in Fig. \ref{fig:aggresive_suggestion} and the orange one corresponds to that in Fig. \ref{fig:conservative_suggestion}. (b) Human suggests an aggressive allocation. The robots are in magenta and the targets are in blue. (c) Human suggests a conservative allocation.}
    \label{fig:IRMRTA_qualitative_example}
\end{figure}

An illustrative example is shown in Fig. \ref{fig:IRMRTA_qualitative_example}. There are ten robots of different sizes (magenta circles) and four targets (blue circles). Let us first look at Fig. \ref{fig:aggresive_suggestion}. Suppose the decision-making algorithm first uses a rational model to find task allocations, i.e., $w(p)=p$ (dotted black line in Fig. \ref{fig:behavior_risk_model}) and $\alpha=1, \beta=1, \delta=0.8$. The resulting allocations are black edges in Fig. \ref{fig:aggresive_suggestion}. In such a case, if the human suggests the allocations should be those blue edges in Fig. \ref{fig:aggresive_suggestion}, it implies that the human perceives the uncertainty in a non-rational way. By solving the IR-MRTA problem, we can find that the human-suggested solution corresponds to parameters $\alpha=0.49, \beta=0.36, \delta=0.75$. The corresponding curve (blue) is shown in Fig. \ref{fig:behavior_risk_model}. From this curve, we can find that the majority of the curve is above the dotted line, i.e., $w(p)>p$, which implies that the human tends to overestimate the success probability in that part. By contrast, if the human suggests the allocations should be those orange edges in Fig. \ref{fig:conservative_suggestion},  we can find the suggestion of the human corresponds to parameters $\alpha=0.75, \beta=1.0, \delta=0.8$. From the associated curve (orange) in Fig. \ref{fig:conservative_suggestion}, we can see that in the part $p>0.4$, the curve is below the dotted line, i.e., $w(p)<p$, which implies the human underestimates the success probability.  Reward and probability parameters used in this example are given in the Appendix. 

\subsection{Quantitative Results}
\subsubsection{Optimality} 
\begin{figure}[t]
    \centering
    \subfloat[]{
    \includegraphics[width=0.23 \textwidth]{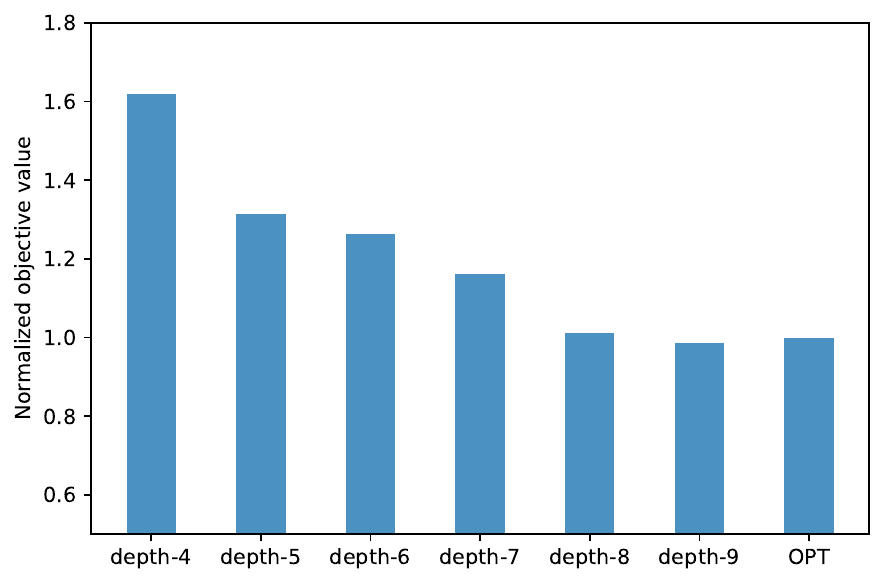}
    \label{fig:optimality}
    } 
    \subfloat[]{
    \includegraphics[width=0.23\textwidth]{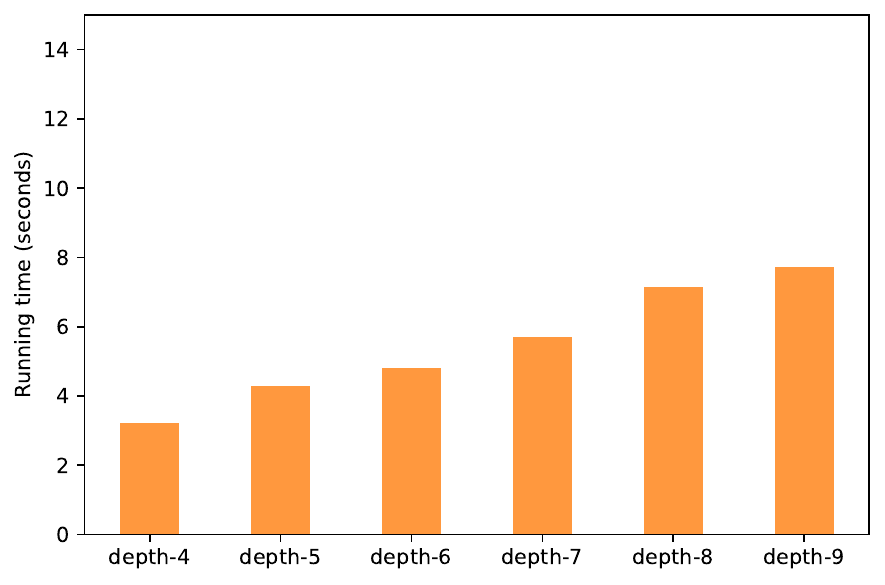}
    \label{fig:running_time_depth}
    }
    \subfloat[]{
    \includegraphics[width=0.23\textwidth]{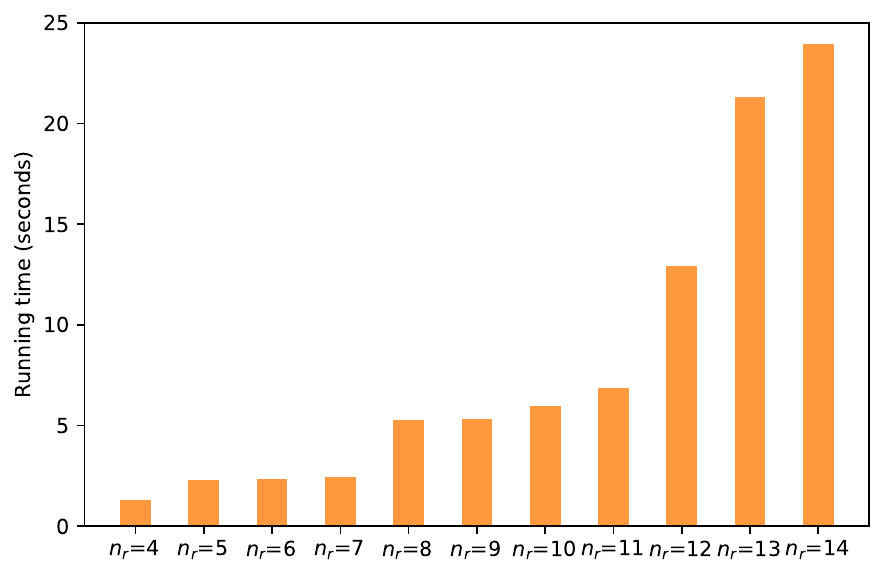}
    \label{fig:running_time_num_robot}
    }
     \subfloat[]{
    \includegraphics[width=0.23\textwidth]{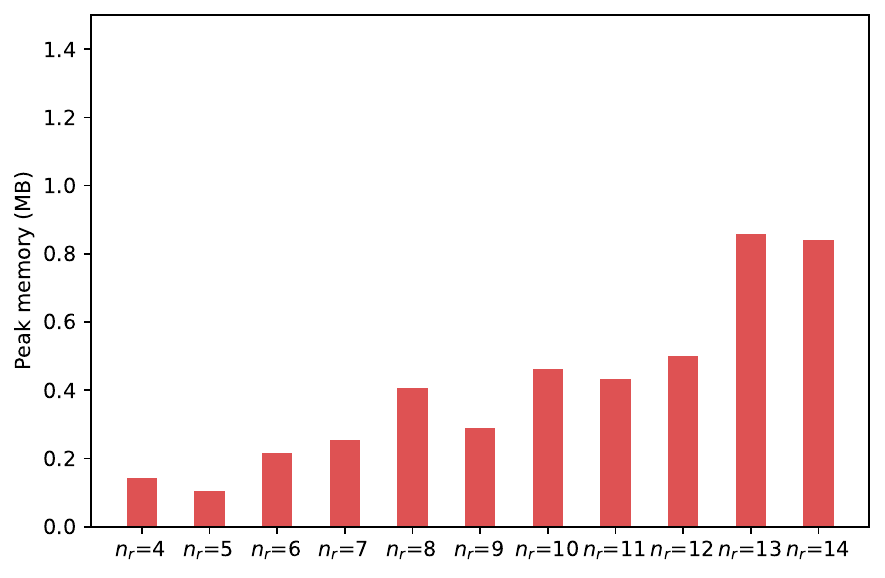}
    \label{fig:memory_peak}
    }
    \caption{
     (a) Optimality. Depth-$x$ denotes the search depth of the subroutine BB-O-IR-MRTA. (b) Running time w.r.t. the search depth. (c) Running time w.r.t. the number of robots. (d) Peak memory usage.}
    \label{fig:quantitative_results}
\end{figure}

We compare the proposed algorithm with the brute-force approach, i.e., directly discretize parameters and enumerate all the possible combinations. As stated in the Theorem \ref{theorem:BB_nonconvex}, the performance of the proposed algorithm depends on the depth of the search tree. We denote the BB-IR-MRTA with a depth $x$ as depth-$x$ in the experiments. We randomly generate a collection of forward problem instances, each of which is associated with a particular $[\alpha, \beta, \delta]$. We set the number of robots and targets to eight for each instance. For each instance, we randomly generate several feasible pseudo-human suggestions (i.e., there is a solution to the inverse problem) and solve the IR-MRTA to obtain $[\hat{\alpha}, \hat{\beta}, \hat{\delta}]$. Let $g^*$ be the optimal objective value obtained using the brute-force approach and $g(sol)$ be the one obtained using BB-IR-MRTA. To statistically compare results across all the instances, we use the normalized objective $\frac{g(sol)}{g^*}$ as the criterion in experiments. The algorithm should make this criterion close to 1. As shown in Fig. \ref{fig:optimality},  as the search depth increases the normalized objective gradually approaches one, which suggests the solution obtained using BB-IR-MRTA is close to the optimal solution (after increasing the depth to eight). 

\subsubsection{Running Time and Peak Memory Usage} We study the running time of the proposed algorithm in two directions. One is to study how the running time changes when the search depth increases. The other is to study how the running time changes as the size of the human suggestion (relates to the size of robots and targets) increases which will affect the size of the search tree in BB-IR-MRTA. In both cases, we set the number of robots to be equal to that of targets, and in Fig. \ref{fig:running_time_depth} the number is set to be eight. Fig. \ref{fig:running_time_depth} and Fig. \ref{fig:running_time_num_robot}, we can see that as the search depth (the number of robots) increases the running time will also increase but not exponentially. Such a mild increase in running time may be attributed to the pruning steps in the algorithm and the algorithm does not need to enumerate all the possible cases.  By contrast, if we use brute-force approaches to solve the problem, the running time can vary between 5 minutes to 10 minutes which depends on the discretizing resolution.  The results of peak memory usage are shown in Fig. \ref{fig:memory_peak}. We can observe that the memory usage will increase mildly as the search depth increases and the overall memory usage is pretty low ($\leq$ 1 MB).

\section{Conclusion}
We introduce a new type of multi-robot task allocation problem named IR-MRTA to accommodate human suggestions in a risk-sensitive multi-robot task allocation setup. We introduce a new branch and bound algorithm to solve IR-MRTA. We validate our formulation and algorithms through numerical simulations. In future work, we plan to address several directions. First, if a human gives multiple suggestions at the same time, how to deal with such suggestions? Second, in this paper, we assume that it is always feasible to solve the inverse problem using human suggestion. In the future, we plan to deal with the case where this is infeasible. Also of interest is the case where the forward algorithm is not greedy - how should we design algorithms to solve inverse problems in this setting?

\bibliographystyle{splncs04}
\bibliography{WAFR24}

\newpage
\section{Appendix}
\subsection{Proof of Theorem \ref{theorem:BB_nonconvex}}
\begin{proof}
    Suppose the algorithm terminates when the depth of the search tree is $n_d \geq 2$. In Algorithm \ref{algorithm:BB_nonconvex},  when we generate child nodes for root nodes (tree depth is 1), we split the set for $\hat{\beta}$ as two sets, i.e., $[\hat{\beta}_{min}, \beta]$ and $[\beta, \hat{\beta}_{max}]$. Similarly, the feasible set for $\hat{\delta}$ is split as two sets, i.e., $[\hat{\delta}_{min}, \delta]$ and $[\delta, \hat{\delta}_{max}]$. After this step (tree depth $\geq 2$), when we generate child nodes for the node $v_p$, we will split the sets stored in $v_p$ evenly for four child nodes, i.e.,  from $[\ubar{\beta}, \bar{\beta}]$ and $[\ubar{\delta}, \bar{\delta}]$ to $[\ubar{\beta}, \frac{\ubar{\beta}+\bar{\beta}}{2}], [\frac{\ubar{\beta}+\bar{\beta}}{2}, \bar{\beta}]$ and $[\ubar{\delta}, \frac{\ubar{\delta}+\bar{\delta}}{2}], [\frac{\ubar{\delta}+\bar{\delta}}{2}, \bar{\delta}]$. As a result, when the search process reaches the depth $n_d$, each leaf node with sets $[\ubar{\beta}, \bar{\beta}]$ and $[\ubar{\delta}, \bar{\delta}]$ should satisfy either $\ubar{\beta} \geq \beta$ ($\ubar{\delta} \geq \delta$) or $\bar{\beta} \leq \beta$ ($\bar{\delta} \leq \delta$). The size of the set satisfies either $\bar{\beta}-\ubar{\beta} = \frac{\beta-\hat{\beta}_{min}}{2^{n_d-1}}$ ($\bar{\delta}-\ubar{\delta} = \frac{\delta-\hat{\delta}_{min}}{2^{n_d-1}}$) or $\bar{\beta}-\ubar{\beta} = \frac{\hat{\beta}_{max}-\beta}{2^{n_d-1}}$($\bar{\delta}-\ubar{\delta} = \frac{\hat{\delta}_{max}-\delta}{2^{n_d-1}}$). 

    Since the leaf nodes include all the feasible combinations of subsets of $\hat{\beta}$ and $\hat{\delta}$, the global minimum, $\hat{\beta}^*$ and $\hat{\delta}^*$, must fall into one of them. Let leaf node $v_{u}$ be the node with the lowest upper bound and $v_{OPT}$ be the node that will generate the global minimum. It should be noted that $v_{OPT}$ is not known to us and it can be any leaf node. Instead, we know two facts. One is that the upper bound of the objective identified in $v_{OPT}$ is higher than that in $v_{u}$, i.e., $v_{u}.UB \leq v_{OPT}.UB$. Another is the optimal objective $g^*$ should be between $v_{OPT}.LB$ and $v_{OPT}.UB$, i.e., $v_{OPT}.LB \leq g^* \leq v_{OPT}.UB$. Therefore, we have
    \begin{equation}\label{eq:bound_obj1}
        v_{u}.UB - g^* \leq v_{u}.UB - v_{OPT}.LB \leq v_{OPT}.UB - v_{OPT}.LB.
    \end{equation}
    Namely, the gap between our best objective identified so far and the optimal objective value is no greater than the gap between the lower bound and upper bound of the leaf node $v_{OPT}$. In the following, we will show how to find the upper bound for $v_{OPT}.UB - v_{OPT}.LB$.

    For any leaf node $v$  with $[\ubar{\beta}, \bar{\beta}]$ and $[\ubar{\delta}, \bar{\delta}]$, there are four cases.
    \subsubsection{Case1}: $\ubar{\beta} \geq \beta$ and $\ubar{\delta} \geq \delta$ In this case, by solving the upper and lower bound formulation, we can find that $v.UB - v.LB=0$

    \subsubsection{Case2}: $\ubar{\beta} \geq \beta$ and $\bar{\delta} \leq \delta$. In this case, the upper and lower bounds can be computed as  
    \begin{align*}
        v.UB = w_{\alpha}\norm{(\alpha^*-\alpha)}+w_{\beta}\norm{(\ubar{\beta}-\beta)}+w_{\delta}\norm{(\ubar{\delta}-\delta)} \\
        v.LB = w_{\alpha}\norm{(\alpha^*-\alpha)}+w_{\beta}\norm{(\ubar{\beta}-\beta)}+w_{\delta}\norm{(\bar{\delta}-\delta)}.
    \end{align*}
   Namely, 
   \begin{equation*}
       v.UB - v.LB = w_{\delta}(\norm{(\ubar{\delta}-\delta)}-\norm{(\bar{\delta}-\delta)}) \leq w_{\delta}\norm{\bar{\delta}-\ubar{\delta}}
   \end{equation*}

    \subsubsection{Case3}:$\bar{\beta} \leq \beta$ and $\ubar{\delta} \geq \delta$. Similar to case 2, we have $v.UB - v.LB \leq w_{\beta}\norm{\bar{\beta}-\ubar{\beta}}$.

     \subsubsection{Case4}: $\bar{\beta} \leq \beta$ and $\bar{\delta} \leq \delta$ Similar to case 2, we have $v.UB - v.LB \leq w_{\beta}\norm{\bar{\beta}-\ubar{\beta}} + w_{\delta}\norm{\bar{\delta}-\ubar{\delta}}$.

     Based on the above discussion, the Eq. \eqref{eq:bound_obj1} can be extended as 
     
     \begin{equation}
         \begin{aligned}
               v_{u}.UB - g^* &\leq v_{OPT}.UB - v_{OPT}.LB \\
               &\leq w_{\beta} \cdot \norm{\bar{\beta}-\ubar{\beta}} + w_{\delta}\norm{\bar{\delta}-\ubar{\delta}} \\
               &\leq w_{\beta} \max(\frac{\beta-\hat{\beta}_{min}}{2^{n_d-1}}, \frac{\hat{\beta}_{max}-\beta}{2^{n_d-1}}) 
               + w_{\delta} \max(\frac{\delta-\hat{\delta}_{min}}{2^{n_d-1}}, \frac{\hat{\delta}_{max}-\delta}{2^{n_d-1}})=\epsilon.
         \end{aligned}
     \end{equation}
\end{proof}

\subsection{Details of Qualitative Example}
The probability and reward matrices for the qualitative example are given below.  In the objective, we set $w_{\alpha}=1, w_{\beta}=1, w_{\delta}=20$.
\[ p=
\begin{bmatrix}
0.85 & 0.74 & 0.59 & 0.41 \\
0.89 & 0.80 & 0.67 & 0.50 \\
0.92 & 0.85 & 0.74 & 0.59 \\
0.94 & 0.89 & 0.80 & 0.67 \\
0.96 & 0.92 & 0.85 & 0.74 \\
0.97 & 0.94 & 0.89 & 0.80 \\
0.98 & 0.96 & 0.92 & 0.85 \\
0.99 & 0.97 & 0.94 & 0.89 \\
0.99 & 0.98 & 0.96 & 0.92 \\
0.99 & 0.99 & 0.97 & 0.94 \\
\end{bmatrix},
r=
\begin{bmatrix}
67.00 & 207.59 & 347.96 & 488.23 \\
67.15 & 207.84 & 348.29 & 488.55 \\
67.26 & 208.05 & 348.60 & 488.91 \\
67.34 & 208.21 & 348.87 & 489.27 \\
67.40 & 208.34 & 349.10 & 489.60 \\
67.43 & 208.43 & 349.28 & 489.90 \\
67.45 & 208.49 & 349.42 & 490.15 \\
67.45 & 208.53 & 349.51 & 490.34 \\
67.45 & 208.55 & 349.58 & 490.49 \\
67.43 & 208.55 & 349.62 & 490.60 \\
\end{bmatrix}
\]
\end{document}